\documentclass[runningheads]{llncs}

\usepackage{eccv}

\usepackage{eccvabbrv}

\renewcommand{\paragraph}[1]{\vspace{.5em}\noindent\textbf{#1.}}

\usepackage{xspace}
\newcommand{\paper}{OmniPoint\xspace}

\usepackage{booktabs} %
\usepackage{amsmath}  %
\usepackage{amssymb}  %
\usepackage{xcolor}   %
\usepackage{graphicx} %
\usepackage{multirow} %
\usepackage[table, xcdraw]{xcolor}

\usepackage[normalem]{ulem}

\DeclareRobustCommand{\bestcap}[1]{\colorbox{ForestGreen!20}{\textbf{#1}}}
\DeclareRobustCommand{\secondcap}[1]{\colorbox{LimeGreen!20}{\uline{#1}}}
\DeclareRobustCommand{\thirdcap}[1]{\colorbox{yellow!20}{#1}}

\usepackage{graphicx}
\usepackage{booktabs}

\usepackage[accsupp]{axessibility}  %

\usepackage[pagebackref,breaklinks,colorlinks,allcolors=eccvblue]{hyperref}

\usepackage{orcidlink}

\begin{document}

\title{OmniPoint: Universal Monocular Metric Pointcloud from Any Camera}

\titlerunning{OmniPoint: Universal Monocular Metric Pointcloud from Any Camera}

\author{
    Botao Ye$^{1,2}$\thanks{Work done as an intern at Google DeepMind.} \and
    Marc Pollefeys$^{2}$ \and
    Ming-Hsuan Yang$^{1}$ \and
    Abhijit Kundu$^{1}$
}

\authorrunning{B.~Ye et al.}

\institute{
    $^1$Google DeepMind \quad $^2$ETH Zurich
}

\maketitle

\begin{center}
    \textbf{Project Page:} \url{https://botaoye.github.io/omnipoint/}
\end{center}

\begin{center}
    \centering
    \includegraphics[width=1.0\textwidth]{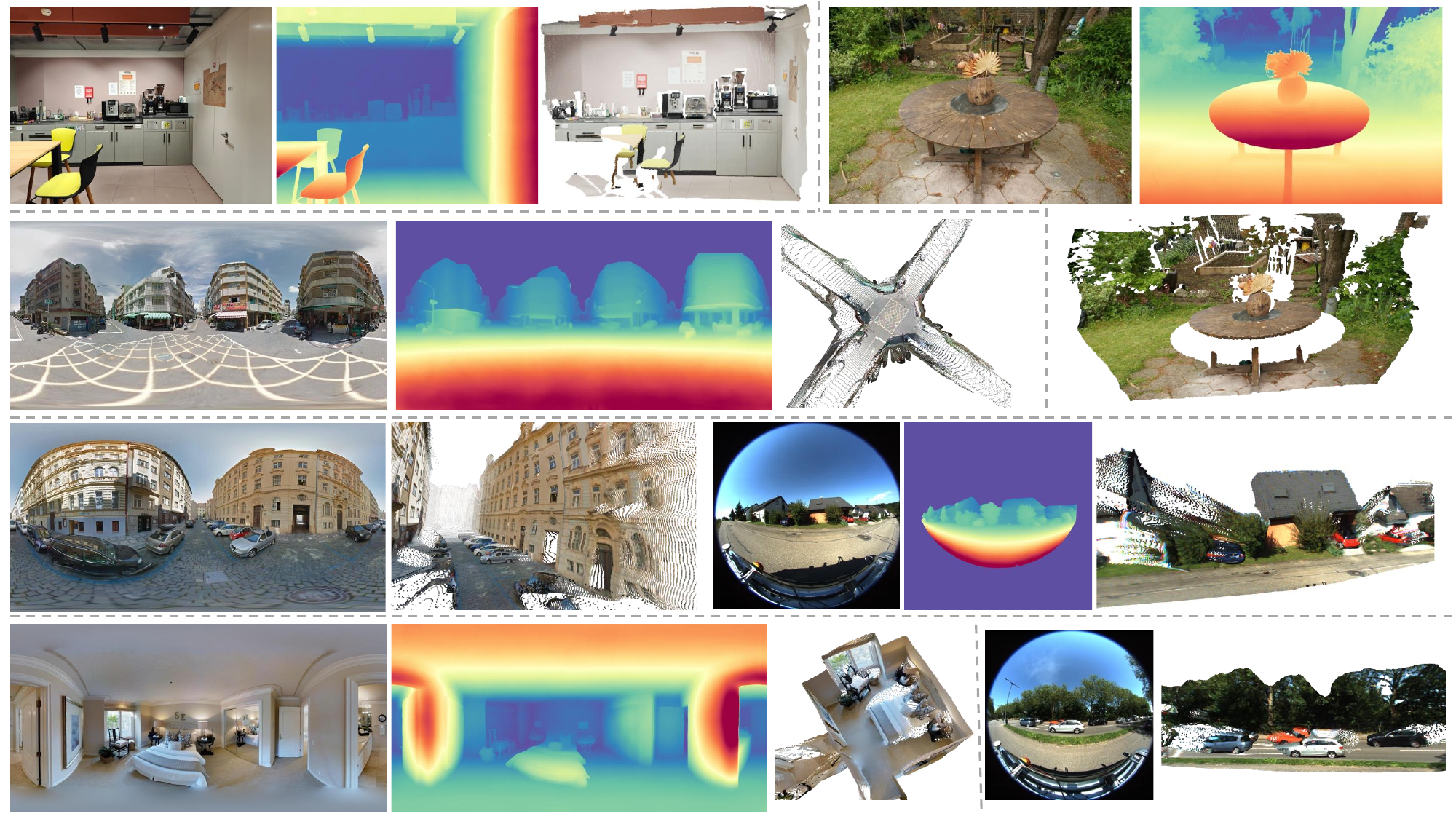}
    \captionof{figure}{\textbf{Universal Monocular Point Cloud Estimation.} 
    \paper{} reconstructs metric 3D point clouds from a single image using a unified framework that supports diverse camera models, including perspective, fisheye, and 360° panoramic images. 
    It further accommodates flexible geometric conditioning by incorporating camera intrinsics and sparse depth information when available.}
    \label{fig:teaser}
\end{center}

\begin{abstract}
    
    Recovering metric 3D geometry from monocular images is a fundamental computer vision task, yet current methods remain heavily fragmented by fixed camera model assumptions and inflexible input schemes. We present \paper{}, a unified framework designed to generalize metric reconstruction across diverse imaging sensors, including pinhole, fisheye, and equirectangular projections, while accommodating varying geometric priors. To overcome projection rigidity, \paper{} abandons conventional planar depth regression. It instead adopts a decoupled ray and distance representation alongside a decoupled training objective, explicitly separating the camera projection model from the scene structure. To address the severe scarcity of training data for alternative cameras, we introduce a bidirectional augmentation strategy that explicitly bridges labeled perspective data and unlabeled omnidirectional domains in 3D space. Furthermore, to seamlessly integrate optional inputs like camera intrinsics or sparse depth without destabilizing the network through feature distribution shifts, we propose a robust information injection mechanism. This mechanism utilizes learnable input state embeddings to resolve architectural ambiguity and applies vectorized Gaussian smoothing to densify irregular measurements. Extensive experiments demonstrate that \paper{} achieves state-of-the-art zero-shot performance across multiple benchmarks, establishing a robust new standard for unified monocular 3D reconstruction.

\end{abstract}

\section{Introduction}
\label{sec:intro}
Understanding the dense 3D structure of a scene from a single monocular image is a cornerstone challenge in computer vision. Recently, the field has witnessed a paradigm shift driven by large-scale datasets~\cite{hypersim, scannetpp} and visual foundation models~\cite{dinov2, stable-diffusion}. By learning powerful priors, modern approaches can now recover high-quality depth maps~\cite{marigold, da} and pixel-wise 3D point clouds~\cite{moge, unidepth} with remarkable zero-shot generalization across diverse, in-the-wild environments.

Despite this progress, current methodologies remain highly fragmented. Existing methods are typically \textit{tied to a fixed camera model}, assuming a single projection model, most commonly pinhole~\cite{da, moge}, or in specialized cases panoramic~\cite{DA2} or fisheye~\cite{xie2023omnividar}. Furthermore, they are \textit{inflexible with respect to input modalities} and cannot dynamically incorporate optional geometric cues such as camera intrinsics or sparse depth measurements (\eg, from LiDAR or SfM).
This fragmentation poses a significant barrier in practical applications such as robotics or autonomous driving, which are frequently equipped with heterogeneous sensor suites (\eg, a forward-facing pinhole camera paired with surround-view fisheye lenses) and varying depth sensors. Deploying a robust 3D perception system in such scenarios currently requires maintaining multiple isolated models, which incurs prohibitive memory overhead and complicates sensor fusion.

To overcome this fragmentation, we introduce \paper{}, a unified framework for monocular geometry estimation centered around a \textit{camera-agnostic 3D representation}. Conventional approaches predict per-pixel depth values $z$ under a pinhole assumption, making their output inherently tied to a specific camera model. For instance, affine-invariant disparity ($1/z$)~\cite{da, dav2} or log-depth ($\log z$)~\cite{eigen2014depth, moge} require $z>0$, precluding $360^\circ$ environments and points behind the camera. More importantly, the projection formula $\mathbf{P} = z \cdot K^{-1}[u, v, 1]^{\top}$ assumes a linear mapping from pixel coordinates to camera rays, which fails for wide-FoV or non-pinhole cameras. While directly predicting per-pixel 3D coordinates $(x, y, z)$ avoids this formula, it \emph{entangles} camera projection with scene geometry: the network must implicitly learn the entire projection model for each camera type, making cross-camera generalization fundamentally difficult. Our key insight is to explicitly \textit{decouple} these two sources of variation. \paper{} adopts a universal representation based on a ray direction $\mathbf{r}$ and a radial distance $d$ for each pixel, defining a 3D point as $\mathbf{P} = d \cdot \mathbf{r}$. The ray direction $\mathbf{r}$ isolates the camera-dependent projection geometry, while the radial distance $d$ encodes the camera-agnostic scene structure along that ray. This separation ensures that once the ray direction is known, the exact same distance prediction applies regardless of the camera type. Furthermore, we pair this representation with a \emph{decoupled training objective} (Sec.~\ref{sec:decoupled_loss}) that prevents ray and distance errors from interfering during optimization, a subtle but critical design choice validated by our ablations (Tab.~\ref{tab:ablation_loss}).

However, a universal representation alone is insufficient. A major bottleneck in achieving true cross-camera generalization is the severe scarcity of non-pinhole training data. To bridge this gap, we propose a systematic \emph{bidirectional data augmentation} strategy. Rather than relying on simple 2D image transformations, we explicitly bridge labeled perspective data and unlabeled omnidirectional data in 3D. A \emph{Perspective-to-Any} process synthetically projects large-scale pinhole datasets into fisheye and panoramic views to generate rich, paired supervision. Conversely, an \emph{Any-to-Perspective} process samples virtual pinhole views from unlabeled in-the-wild panoramic images, generates pseudo-ground truth using our perspective model, and reprojects them to enforce cross-camera consistency.

Beyond camera generalization, another critical capability for a unified system is the flexible integration of geometric priors. A major obstacle in prior-aware models is that toggling optional inputs (\eg, switching from RGB-only to RGB+sparse depth) causes severe feature distribution shifts, which destabilizes network training. Our insight is to explicitly signal the active configuration using learnable \textit{input-state embeddings}, which dynamically guides the Vision Transformer (ViT) to resolve architectural ambiguity and correctly interpret varying input distributions. Furthermore, to effectively utilize sparse depth without suffering from its irregular and noisy nature, we introduce a robust information injection mechanism via fully vectorized Gaussian smoothing. Instead of feeding raw points directly, this mechanism normalizes and densifies the measurements into spatially consistent geometric guidance. This allows \paper{} to seamlessly transition between acting as a zero-shot monocular estimator and a high-performance depth densification system.

\begin{table*}[t]
  \centering
  
\caption{Comparison of \paper{} with state-of-the-art monocular geometry estimation methods. 
\paper{} is the first approach to unify all camera types, output representations, and geometric input priors within a single framework.}
  \label{tab:comparison}
  \small
  \resizebox{\textwidth}{!}{%
  \begin{tabular}{ll cccccccccc}
    \toprule
    & & DAv2~\cite{dav2} & MoGe~\cite{moge} & MoGeV2~\cite{mogev2} & DepthPro~\cite{depth_pro} & UniK3D~\cite{unik3d} & DAC~\cite{dac} & DA$^2$~\cite{DA2} & PromptDA~\cite{promptDA} & PriorDA~\cite{priorDA} & \textbf{OmniPoint} \\
    \midrule
    \multirow{3}{*}{\textbf{Output}} & Affine-Inv Depth & \checkmark & \checkmark & \checkmark & \checkmark & \checkmark & \checkmark & & \checkmark & \checkmark & \checkmark \\
    & Affine-Inv Points & & \checkmark & \checkmark & \checkmark & \checkmark & & & & & \checkmark \\
    & Metric Points & & & \checkmark & \checkmark & \checkmark & & & & & \checkmark \\
    \midrule
    \multirow{3}{*}{\shortstack{\textbf{Image}\\\textbf{Type}}} & Pinhole & \checkmark & \checkmark & \checkmark & \checkmark & \checkmark & \checkmark &  & \checkmark & \checkmark & \checkmark \\
    & Fisheye & & & & & \checkmark & \checkmark & & & & \checkmark \\
    & Panorama & & & & & \checkmark & \checkmark & \checkmark & & & \checkmark \\
    \midrule
    \multirow{2}{*}{\textbf{Prior}} & Intrinsic & & & & & \checkmark & & & & & \checkmark \\
    & Sparse Depth & & & & & & & & \checkmark & \checkmark & \checkmark \\
    \bottomrule
  \end{tabular}
  }
  \vspace{-3mm}
\end{table*}

Our key contributions are:
\begin{itemize}
\item \paper{}, the first unified monocular geometry estimation framework that effectively handles heterogeneous camera models (pinhole, fisheye, 360$^\circ$) and flexible priors (intrinsics, sparse depth) without architectural changes.
\item A bidirectional augmentation strategy that effectively bridges the supervision gap between perspective and omnidirectional domains.
\item A robust geometric information injection mechanism that prevents feature distribution shifts and leverages sparse inputs for highly accurate metric prediction.
\item State-of-the-art zero-shot performance across a wide range of camera models and benchmarks, demonstrating unparalleled flexibility.
\end{itemize}

\section{Related Work}
\label{sec:related}

\paragraph{Monocular Depth Estimation}
Early approaches to monocular depth estimation relied on CNNs trained with supervised or self-supervised signals from stereo pairs or monocular video. A recent paradigm shift has been driven by models trained on large-scale, diverse datasets~\cite{midas, da, moge} and by pre-trained foundation models~\cite{marigold, dav2}. These methods demonstrate strong zero-shot generalization to in-the-wild images. However, a fundamental limitation remains: they are designed for and evaluated primarily on perspective images under the pinhole camera assumption. Their planar-depth outputs are consequently ill-suited to other camera geometries.

\paragraph{Omnidirectional and Wide-FoV Depth}
Another line of work targets specialized, non-pinhole camera systems. Methods have been proposed specifically for 360$^\circ$ panoramas~\cite{panda, DA2} and for fisheye or heavily distorted lenses~\cite{lee2023slabins, simfir, xie2023omnividar}, typically relying on camera-specific unprojection models. While effective within their respective domains, these models are highly specialized: a model trained for panoramas cannot process fisheye images, and neither can it operate on standard pinhole images, limiting their applicability in general scenarios.
UniK3D~\cite{unik3d} takes a step toward unifying multiple camera models within a single framework. However, due to the scarcity of wide-FoV and non-pinhole training data, its performance on panoramic and fisheye images remains suboptimal. In contrast, we show that our approach significantly outperforms UniK3D in these settings (see Fig.~\ref{fig:compare1} and Tab.~\ref{tab:comparison}), owing to our bidirectional data augmentation strategy. Moreover, UniK3D is not designed to incorporate geometric prior information such as sparse depth or intrinsics.

\paragraph{Conditional Geometry Estimation}
A third line of research incorporates available depth information into the estimation process. The most prominent examples are depth completion methods~\cite{ogni, tang2024bilateral}, which densify sparse depth inputs (e.g., from LiDAR). More recent promptable approaches~\cite{promptDA, priorDA} guide a general depth estimation model using sparse points or other auxiliary cues. However, these methods still rely on the pinhole camera assumption and cannot accommodate diverse camera models or leverage camera intrinsics as conditions. In addition, they are tailored for conditioned estimation only and cannot be applied to non-conditioned RGB inputs.

\paper{} is the first work to bridge these three previously disjoint research directions. It provides a single, unified model that functions simultaneously as (1) a general-purpose pinhole depth estimator, (2) a specialized wide-FoV estimator, and (3) a conditional, prior-aware model—without any architectural changes.

\section{Method}
\label{sec:method}

\paper{} is a feedforward framework that predicts dense 3D point clouds from a single monocular image captured by any camera type. Given an input image $\mathbf{I}$ and optional geometric priors such as camera intrinsics $K$ and sparse depth $\mathbf{S}$, the model outputs a 3D point $\mathbf{P}_i$ for each pixel $i$, represented by its ray direction and radial distance from the camera center:
\begin{equation}
    f_{\text{OmniPoint}}(\mathbf{I}, [K, \mathbf{S}]) = \{\hat{s}, \hat{\mathbf{R}}, \hat{\mathbf{D}}\},
\end{equation}
where $\tilde{\mathbf{R}}$ denotes the predicted ray directions, $\hat{\mathbf{D}}$ the corresponding radial distances, and $s$ is the global metric scale factor. The final metric point map may be obtained via $\hat{s}\cdot\hat{\mathbf{R}} \cdot \hat{\mathbf{D}}$.

\subsection{Camera-Agnostic 3D Representation}
\label{sec:representation}
A core design choice in a camera-universal model is the output representation. We analyze common formulations and motivate our unified one.

\paragraph{Planar Depth}
The standard paradigm in monocular depth estimation~\cite{marigold, dav2} is to regress planar depth $z$ orthogonal to the camera plane. This formulation inherently relies on the linear unprojection formula $\mathbf{P} = z \cdot K^{-1}[u, v, 1]^{\top}$. This mapping fundamentally breaks down for wide-FoV or non-pinhole cameras, where rays undergo complex radial distortions and do not project linearly. 
Moreover, planar depth cannot represent full $360^\circ$ panoramic scenes. As the field of view approaches $180^\circ$, $z$ diverges to infinity, and it becomes undefined for points located behind the camera ($z < 0$). Therefore, common prediction targets such as affine-invariant disparity ($1/z$)~\cite{da, dav2} and log-depth ($\log z$)~\cite{eigen2014depth, moge} inherently require $z > 0$, making them unsuitable for modeling environments that extend beyond the front-facing half-space.
Although some methods, such as UniDepth~\cite{unidepth}, additionally estimate camera rays, they still fall into this category as their primary prediction target remains planar depth.

\paragraph{Pixel-wise 3D Coordinates}
An alternative approach~\cite{dust3r, moge, vggt} directly predicts $(x, y, z)$ coordinates for each pixel.
However, this is fundamentally \textit{camera-dependent} and \textit{ill-conditioned for generalization}, and merely shifts the burden entirely onto the network's latent capacity. The model is forced to memorize a highly non-linear mapping $\mathcal{F}: (u,v) \mapsto (x,y,z)$ that inherently mixes the scene's structural distance with the specific camera's intrinsic projection pattern. Training a single neural architecture to dynamically alternate between the fundamentally different $\mathcal{F}$ mappings of pinhole, fisheye, and equirectangular images leads to optimization conflicts and poor zero-shot generalization across cameras.

\paragraph{Ray and Distance}
To achieve true camera-agnostic estimation, we explicitly factorize the prediction into a unit ray direction $\mathbf{r} \in \mathbb{R}^3$ and a scalar radial distance $d \in \mathbb{R}^+$. The ray $\mathbf{r}$ perfectly absorbs any arbitrary lens distortion or projection geometry, leaving $d$ to represent pure structural distance. By doing so, the distance prediction becomes completely invariant to the camera model. Furthermore, while recent camera-universal works like UniK3D~\cite{unik3d} utilize Spherical Harmonics (SH) to parameterize the ray field, we find that SH introduces unwanted smoothing artifacts and struggles to capture abrupt geometric changes in highly distorted edge regions (See Fig.~\ref{fig:compare1} and Fig.~\ref{fig:compare2}). Our simple, explicit ray-distance pairing performs comparably or better, providing a stable, high-fidelity target for network regression.

\begin{figure*}[t]
    \centering
    \includegraphics[width=\linewidth, trim={0mm 0cm 0 0cm},clip]{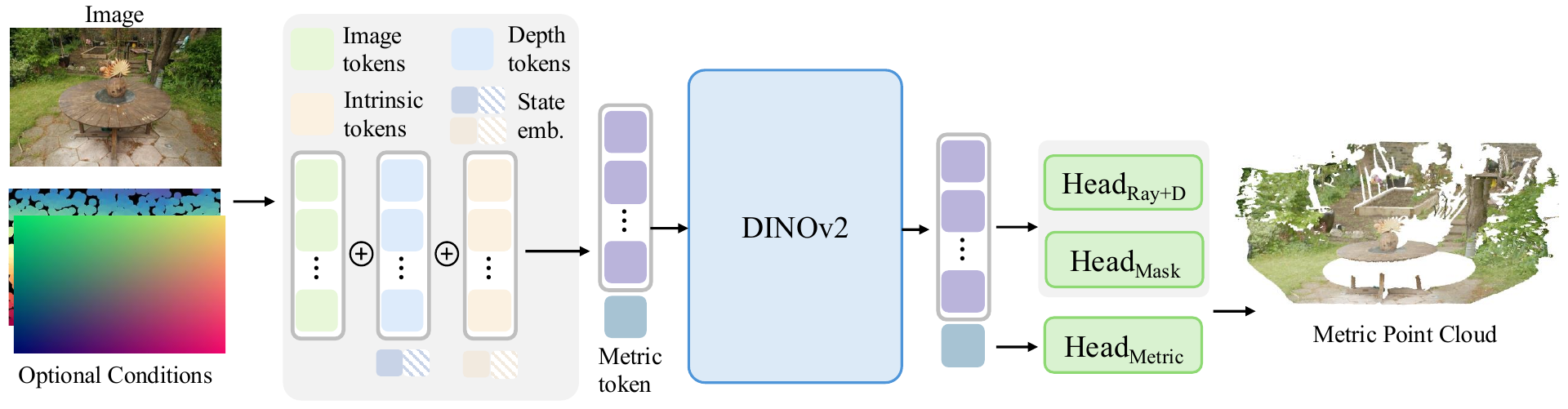}
    \caption{\textbf{Overview of \paper{}.} Our pipeline takes a single image from any camera and optional conditions (intrinsics, sparse depth). It uses a universal ray-distance representation to produce a metric point cloud.} 
    \label{fig:pipeline}
\end{figure*}

\begin{figure*}[t]
    \centering
    \includegraphics[width=\linewidth, trim={0mm 0cm 0 0cm},clip]{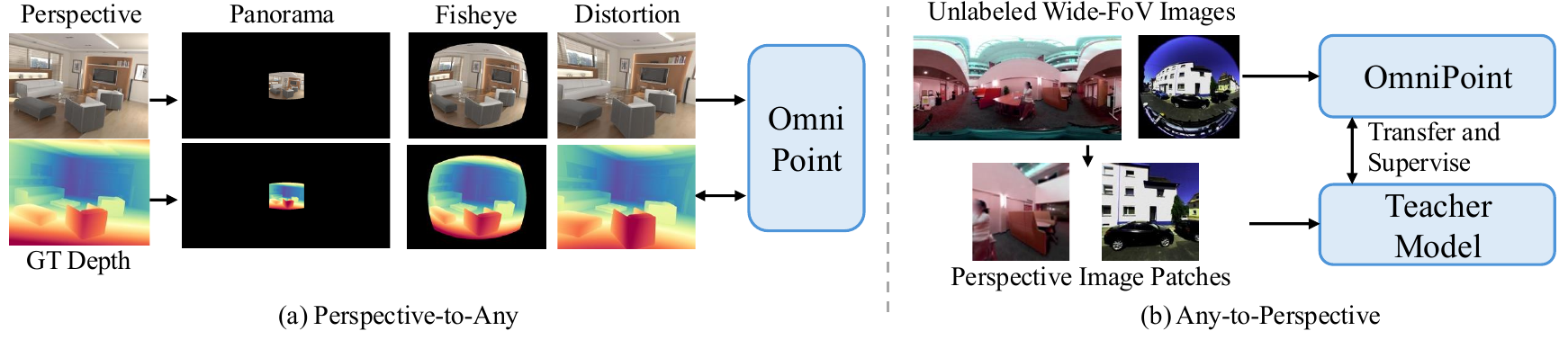}
    \caption{\textbf{Bidirectional Data Augmentation.} (a) \textbf{Perspective-to-Any:} We synthesize novel fisheye and panoramic views from pinhole datasets. (b) \textbf{Any-to-Perspective:} We leverage unlabeled wide-FoV data by reprojecting virtual pinhole views for self-training.} 
    \label{fig:bidirec_aug}
    \vspace{-3mm}
\end{figure*}

\subsection{Decoupled Point Learning}
\label{sec:decoupled_loss}

Predicting both distance and ray direction introduces competing gradients if supervised naively. A standard $L_1$ point loss, $\mathcal{L}_{\text{naive}} = \sum_i \|\hat{d}_i \cdot \hat{\mathbf{r}}_i - d_i \cdot \mathbf{r}_i\|_1$, entangles the errors of the two predictions. For instance, if the network predicts an incorrect ray direction $\hat{\mathbf{r}}_i$ but the correct distance $\hat{d}_i$, the resulting 3D point $\hat{\mathbf{P}}_i$ will still deviate from the ground truth. Consequently, the gradient update will inappropriately penalize the accurate distance prediction to compensate for the ray error. This cross-contamination destabilizes the learning process, particularly in highly distorted regions like the periphery of fisheye lenses.

To prevent this interference, we introduce a decoupled training objective that supervises the camera geometry and scene structure independently:
\begin{equation}
\mathcal{L}_{\text{ray}} = \sum_i \|\hat{\mathbf{r}}_i - \mathbf{r}_i\|_1, \quad
\mathcal{L}_{\text{point}} = \sum_i \|s^{*} \cdot \hat{d}_i \cdot \mathbf{r}_i - d_i \cdot \mathbf{r}_i\|_1,
\end{equation}
where $s^{*}$ denotes the optimal scale obtained online via ROE alignment~\cite{moge} between the predicted affine-invariant point map ($\hat{d}_i \cdot \mathbf{r}_i$) and the ground-truth point map.
In $\mathcal{L}_{\text{point}}$, we construct a proxy 3D point by projecting the \emph{predicted} distance $\hat{d}_i$ along the \emph{ground-truth} ray $\mathbf{r}_i$. This cleanly isolates the distance error, ensuring that the depth estimator is penalized strictly for structural inaccuracies, while $\mathcal{L}_{\text{ray}}$ independently guides the network to learn the correct unprojection mapping. As demonstrated in our ablations (Tab.~\ref{tab:ablation_loss}), this decoupling is crucial for achieving high-fidelity point clouds across diverse camera geometries.

\subsection{Bidirectional Data Augmentation}
\label{sec:augmentation}
To overcome the severe scarcity of ground-truth 3D data for non-pinhole cameras, we introduce a systematic bidirectional data augmentation pipeline. This strategy explicitly bridges the gap between labeled perspective data and unlabeled omnidirectional domains operating in 3D space.

\paragraph{Perspective-to-Any Synthesis}
To generate rich, paired supervision for diverse camera models, we simulate wide-FoV cameras from existing large-scale pinhole datasets~\cite{hypersim, vkitti, scannetpp, hoi4d}. Given a pinhole image $I_p$ and its ground-truth depth $D_p$, we first unproject them into a dense 3D point cloud $\mathbf{P}$. We then define a virtual camera model $f_{\text{any}}$ (e.g., a fisheye lens with specific radial distortion or an equirectangular panorama) and re-project $\mathbf{P}$ onto this new image plane. We apply a binary validity mask during re-projection to exclude void holes caused by occlusions, thereby creating a highly accurate synthetic training pair $(I_{\text{any}}, (\mathbf{R}_{\text{any}}, D_{\text{any}}))$.

\paragraph{Any-to-Perspective Self-Training}
Conversely, to leverage the vast amount of unlabeled wide-FoV data available in-the-wild, we propose a self-training augmentation. Given an unlabeled fisheye or panoramic image $I_{\text{any}}$, we first sample one perspective patch $I_p$ by re-projecting portions of the image using a virtual pinhole camera. We run our (already strong) pinhole-capable model to get pseudo-ground-truth depth $D_p$ for the sampled patch.
Notably, although we can sample multiple-perspective images from the wide-FoV images, such as panoramic images, we always only sample one path per sample during training, as it avoids inconsistent depth prediction between the predictions of different patches.
These depths are then unprojected and stitched back into the original $I_{\text{img}}$ coordinate frame, creating a new pseudo-label $D_{\text{any}}$ for fine-tuning.

\subsection{Optional Geometric Inputs}
\label{sec:geo_condition}

A critical capability for a unified geometry model is the flexible integration of auxiliary priors, such as camera intrinsics or sparse depth measurements. However, toggling these inputs causes severe feature distribution shifts that can destabilize the network. We address this through a robust information injection mechanism.

\paragraph{Input-State Embeddings}
To resolve the architectural ambiguity caused by varying input configurations (\eg, RGB-only vs. RGB + sparse depth), we explicitly signal the active configuration to the model. We introduce learnable \textit{input-state embeddings} ($\mathbf{e}_{\text{intrinsic}}$ and $\mathbf{e}_{\text{depth}}$) that are added to the token sequence entering the Vision Transformer (ViT). Each embedding acts as a boolean indicator, adopting one learned vector when its respective prior is available and another when it is absent. This enables the backbone to dynamically correctly interpret the incoming feature distributions.

\paragraph{Robust Geometric Injection}
When camera intrinsics are available, we precompute the ground-truth per-pixel ray map $\mathbf{R}_{\text{gt}}$, process it through a lightweight convolutional encoder, and fuse it with the image features. 

When sparse depth $\mathbf{S}$ (\eg, from LiDAR or SfM) is provided, feeding these raw, irregular points directly into the network often leads to noisy gradients.
Instead of directly feeding raw sparse depth as input, we first normalize it by the mean of all valid depth values to ensure stable scaling and consistent depth ranges across different input conditions. Rather than introducing additional scale factors for metric depths as input, we perform a simple post-alignment between the sparse condition and the predicted depth when metric depth is provided. 
This works well because our model can sufficiently utilize normalized input, thus providing correspondence between output and input condition.
Moreover, raw sparse depth $\mathbf{S}$ (from LiDAR, SfM, etc.) provides strong geometric cues but is often spatially irregular and noisy. We apply a fully vectorized Gaussian splatting procedure to obtain a smooth and spatially consistent depth map. Each valid sparse point $\mathbf{S}(u,v)$ is treated as the center of a Gaussian kernel whose influence is distributed within a fixed radius $r$. The spatial weight for an offset $(\Delta u, \Delta v)$ is defined as
\begin{equation}
w(\Delta u, \Delta v) = \exp\!\left(-\frac{\Delta u^2 + \Delta v^2}{2\sigma^2}\right),
\end{equation}
where $\sigma$ scales proportionally to the depth magnitude to maintain relative smoothness across varying depths. All pixel-wise contributions are accumulated in parallel using a scatter-add operation:
\begin{equation}
\mathbf{D}_{\text{smooth}}(x,y) = \frac{\sum_i w_i(x,y) \, S_i}{\sum_i w_i(x,y)},
\end{equation}
yielding a dense and spatially consistent smoothed depth map. We additionally construct a binary mask $\mathbf{M}$ indicating whether each pixel corresponds to an original valid sparse measurement.
$\mathbf{D}_{\text{smooth}}$ and $\mathbf{M}$ are concatenated and projected into the feature space using a convolutional layer, then fused with image embeddings to guide the ViT backbone.
This mechanism allows \paper{} to seamlessly transition into a highly accurate depth densification system.

\subsection{Training}

\paragraph{Loss Functions}
In addition to $\mathcal{L}_{\text{point}}$ and $\mathcal{L}_{\text{ray}}$ (Sec.~\ref{sec:decoupled_loss}), we employ a metric alignment loss~\cite{mogev2} to enforce global scale consistency between the predicted and ground-truth point clouds:
\begin{equation}
\mathcal{L}_{\text{metric}} 
= \| \log(\hat{s}) - \text{stopgrad}(\log(s^{*})) \|_2^2.
\end{equation}
Here, $s^{*}$ is the optimal scalar between the predicted point cloud and the ground truth (see Sec.~\ref{sec:decoupled_loss}). The stop-gradient operator prevents gradients from flowing through $s^{*}$, stabilizing training while encouraging the network to recover the correct global scale.
To further improve geometric fidelity, we incorporate normal and local consistency losses ($\mathcal{L}_{\text{normal}}$, $\mathcal{L}_{\text{local}}$) following~\cite{moge}, which enhance surface smoothness and preserve fine-grained structure. In addition, we apply a binary mask loss $\mathcal{L}_{\text{mask}}$ to mask out sky regions.
The final training objective is given by
\begin{equation}
\begin{split}
\mathcal{L} = \mathcal{L}_{\text{point}} 
+ \lambda_{\text{ray}} \mathcal{L}_{\text{ray}} 
+ \lambda_{\text{metric}} \mathcal{L}_{\text{metric}} \\
+ \lambda_{\text{normal}} \mathcal{L}_{\text{normal}} 
+ \lambda_{\text{local}} \mathcal{L}_{\text{local}} 
+ \lambda_{\text{mask}} \mathcal{L}_{\text{mask}},
\end{split}
\end{equation}
where $\lambda_{\star}$ controls the weight of each term.

\paragraph{Training Process}
Training proceeds in three stages.  
(1) We pretrain on large-scale perspective datasets to establish a strong baseline geometry model.  
(2) We fine-tune on mixed synthetic and real data, including fisheye and panoramic images, using the bidirectional augmentation in Sec.~\ref{sec:augmentation}.  
(3) Finally, we freeze the backbone and train only the mask head using $\mathcal{L}_{\text{mask}}$ with pseudo-labels from SegFormer~\cite{segformer}, ensuring robust sky-region masking during inference.

\begin{table*}[t]
\centering
\caption{\textbf{Relative geometric and depth comparison} across small FoV (8 pinhole datasets: NYUv2~\cite{nyuv2}, KITTI~\cite{kitti}, ETH3D~\cite{eth3d}, iBims-1~\cite{ibim}, GSO~\cite{gso}, Sintel~\cite{sintel}, DIODE~\cite{diode}, and HAMMER~\cite{hammer}), large FoV (fisheye dataset: KITTI360~\cite{kitti360}), and 360° images (Stanford2D3D-S~\cite{stanford2d3d} and PanoSUNCG~\cite{panosuncg}). \paper{} achieves SOTA or competitive performance across all camera types (\bestcap{Best}, \secondcap{second-best}, and \thirdcap{third-best}).}
\label{tab:relative_geo}

\resizebox{\textwidth}{!}{
\begin{tabular}{l cc cc cc cc cc cc}
\toprule
& \multicolumn{4}{c}{\textbf{S.FoV}} 
& \multicolumn{4}{c}{\textbf{L.FoV}} 
& \multicolumn{4}{c}{\textbf{360 Images}} \\
\cmidrule(lr){2-5}\cmidrule(lr){6-9}\cmidrule(lr){10-13}
& \multicolumn{2}{c}{Depth} & \multicolumn{2}{c}{Point}
& \multicolumn{2}{c}{Depth} & \multicolumn{2}{c}{Point}
& \multicolumn{2}{c}{Depth} & \multicolumn{2}{c}{Point} \\
\cmidrule(lr){2-3}\cmidrule(lr){4-5}
\cmidrule(lr){6-7}\cmidrule(lr){8-9}
\cmidrule(lr){10-11}\cmidrule(lr){12-13}
Method 
& Rel$\downarrow$ & $\delta\uparrow$ 
& Rel$\downarrow$ & $\delta\uparrow$
& Rel$\downarrow$ & $\delta\uparrow$
& Rel$\downarrow$ & $\delta\uparrow$
& Rel$\downarrow$ & $\delta\uparrow$
& Rel$\downarrow$ & $\delta\uparrow$ \\
\midrule

DA V1        & 6.43 & 95.3 & --   & --   & -- & -- & --   & --   & --   & --   & --   & --   \\
DA V2        & 6.35 & 95.0 & --   & --   & -- & -- & --   & --   & --   & --   & --   & --   \\
Metric3D V2  & 6.06 & 95.3 & --   & --   & --  & -- & --   & --   & --   & --   & --   & --   \\
UniDepth V2  & 4.23 & 96.6 & 6.06 & 95.1 & \secondcap{7.81}  & \secondcap{93.0} & 18.4 & 79.6 & 19.3 & 69.5 & 102.8 & 2.89 \\
Depth Pro    & 5.28 & 95.6 & 7.49 & 93.3 & 26.6  & 54.6 & 36.2 & 35.5 & --   & --   & --    & --   \\
MoGe V1      & \thirdcap{4.05} & 96.6 & \bestcap{5.25} & \secondcap{95.5} & 9.48 & 90.1 & 28.1 & 52.5 & 22.3 & 62.9 & 102.6 & 2.91 \\
MoGe V2      & \bestcap{3.98} & \bestcap{96.8} & \thirdcap{5.59} & \secondcap{95.5} & 12.0 & 85.0 & 23.4 & 60.9 & 25.1 & 56.1 & 102.8 & 2.85 \\
DA3          & 4.78 & 95.5 & 6.33 & 93.7 & 12.8  & 83.4 & 29.9 & 47.5 & 25.5 & 55.7 & 101.5 & 2.75 \\
UniK3D       & 4.14 & \thirdcap{96.7} & 5.61 & \secondcap{95.5} & \thirdcap{8.77} & \thirdcap{92.7} & \secondcap{11.5} & \secondcap{90.6} & \thirdcap{10.4} & \thirdcap{90.0} & \thirdcap{11.4}  & \thirdcap{89.4} \\
DA$^2$       & --   & --   & --   & --   & --    & --   & --   & --   & \secondcap{6.65} & \secondcap{94.7} & \secondcap{7.30} & \secondcap{94.0} \\
\rowcolor{gray!10}
\textbf{Ours}& \secondcap{4.04} & \bestcap{96.8} & \secondcap{5.55} & \bestcap{95.6} & \bestcap{6.37} & \bestcap{94.0} & \bestcap{6.66} & \bestcap{93.8} & \bestcap{5.79} & \bestcap{95.1} & \bestcap{5.90} & \bestcap{95.1} \\
\bottomrule
\end{tabular}
} %
\vspace{-3mm}
\end{table*}

\begin{figure*}[t]
  \centering
  \includegraphics[width=\linewidth, trim={0mm 0cm 0 0cm},clip]{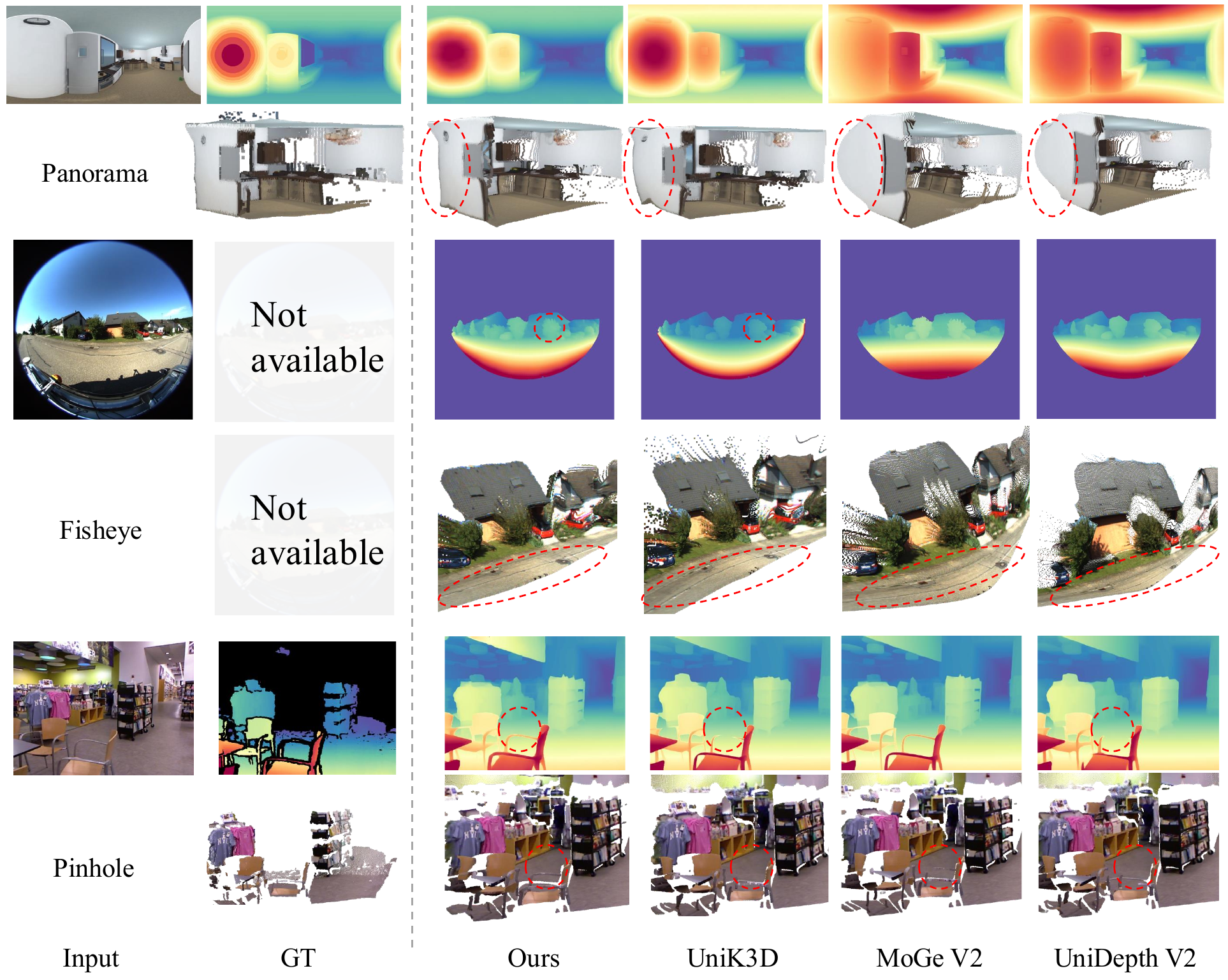}
  \caption{\textbf{Qualitative Comparison on Diverse Camera Models.} We compare \paper{} against baselines on panoramic (top), fisheye (middle), and pinhole (bottom) images. Our method produces geometrically consistent reconstructions, such as the vertical wall in the panorama and the straight road in the fisheye, where UniK3D~\cite{unik3d} shows distortion. \paper{} also recovers finer details in depth maps and point clouds.}
  \label{fig:compare1}
  \vspace{-4mm}
\end{figure*}

\section{Experiments}
\label{sec:exp}

\subsection{Experimental Setup}

\paragraph{Training Datasets}
We employ 29 labeled datasets for training: ARKitScenes~\cite{arkitscenes}, ARKitScenes-HighRes~\cite{arkitscenes}, HOI4D~\cite{hoi4d}, Matterport3D~\cite{matterport3d}, ScanNet++~\cite{scannetpp}, SmartPortraits~\cite{smartportraits}, Waymo~\cite{waymo}, WildRGBD~\cite{wildrgbd}, 3D Ken Burns~\cite{3dkb}, BEDLAM~\cite{bedlam}, BlendedMVS~\cite{blendedmvs}, Dynamic Replica~\cite{dynamicReplica}, EDEN~\cite{eden}, HyperSim~\cite{hypersim}, IRS~\cite{irs}, MVS-Synth~\cite{mvssynth}, OmniObject3D~\cite{omniobject3d}, 
OmniWorld-Game~\cite{omniworld},  Spring~\cite{spring}, Synscapes~\cite{synscapes}, 
PointOdyssey~\cite{pointodyssey},
TartanAir~\cite{tartanair}, TartanAir v2~\cite{tartanair}, Unreal4K~\cite{unreal4k}, UrbanSyn~\cite{urbanSyn}, VirtualKITTI2~\cite{vkitti}, Structure3D~\cite{Structured3D}, and KITTI360~\cite{kitti360}.
We additionally use the unlabeled panorama datasets Diverse360~\cite{diverse360} and 360+x~\cite{chen2024360+} for our Any-to-Perspective training, where KITTI360~\cite{kitti360} is also included.
Please refer to the supplementary material for detailed descriptions of these datasets.

\paragraph{Implementation Details}
We adopt DINOv2-ViT-Large~\cite{dinov2} as the backbone and use DPT~\cite{dpt} heads to predict the 3D points and mask as in~\cite{mogev2}.
For the first training stage, we use 72 A100 GPUs with a batch size of 8 per GPU and train for 10k steps, which takes approximately 20 hours.
The second stage uses the same hardware configuration but reduces the training schedule to 5k steps.
Additional implementation details are provided in the supplementary material.

\paragraph{Baselines}
We compare our approach with the following monocular geometry estimation methods:
1) \emph{Affine-invariant depth estimator}: Depth Anything v1~\cite{da}, Depth Anything v2~\cite{dav2}, MoGe~\cite{moge}, and VGGT~\cite{vggt};
2) \emph{Metric depth estimator}: ZeoDepth~\cite{zoedepth}, Metric3Dv2~\cite{metric3dv2}, UniDepth~\cite{unidepth}, UniDepth v2~\cite{unidepthv2}, Depth Pro~\cite{depth_pro}, MoGe v2~\cite{mogev2}, Depth Anything 3~\cite{da3}, and UniK3D~\cite{unik3d};
3) \emph{Panorama depth estimator}: DA$^2$~\cite{DA2}.

\paragraph{Evaluation Details}
To ensure a strictly fair and rigorous comparison, we re-evaluate all baselines using identical input formats and a standardized metric computation pipeline. We report geometric performance on both depth and 3D point predictions using the standard relative error (Rel$\downarrow$) and the percentage of robust inliers ($\delta_1\uparrow$) following~\cite{moge}.

\begin{table*}[t]
\centering
\caption{\textbf{Metric Depth Estimation ($\delta_1$) Performance.} We evaluate zero-shot metric depth on six diverse benchmarks. Our model outperforms previous SOTA methods. Conditioning on intrinsics (+ K) or sparse depth (+ D) further improves results.}
\vspace{-3mm}
\begin{tabular}{lcccccccc}
\toprule
Method & NYUv2 & KITTI & ETH3D & iBims-1 & DIODE & HAMMER & Mean \\
\midrule
ZoeDepth    &91.9 & 85.4 & 33.7 & 67.2 & 29.3 & 3.23 & 51.8 \\
UniDepth V2 &92.8 & \textbf{95.4} & 69.5 & \textbf{93.2} & 51.8 & 46.8 & 74.4 \\
Depth Pro   &91.9 & 38.3 & 32.8 & 81.5 & 37.7 & 63.0 & 57.5 \\
UniK3D      &94.4 & 93.6 & 83.7 & 92.8 & 73.0 & 58.3 & 82.6 \\
MoGe V2     &\textbf{96.1} & 62.9 & 90.8 & 83.0 & 66.4 & 65.6 & 77.5 \\
\rowcolor[gray]{0.9}
\textbf{Ours}        & 86.2 & 88.0 & \textbf{91.8} & 87.7 & \textbf{73.3} & \textbf{73.8} & \textbf{83.5} \\
\midrule
\rowcolor[gray]{0.9}
Ours + K    & 88.3 & 90.7 & 93.0 & 88.3 & 75.2 & 74.5 & 85.0 \\
\rowcolor[gray]{0.9}
Ours + D    & 98.5 & 98.4 & 94.1 & 99.0 & 98.4 & 99.3 &98.0 \\
\bottomrule
\end{tabular}
\label{tab:metric_depth}
\vspace{-1mm}
\end{table*}

\begin{table}[t]
\centering
\caption{\textbf{Comparison of different input configurations.} Our optional geometric conditions consistently improve relative geometry estimation across metrics.}
\vspace{-2mm}
\begin{tabular*}{0.55\linewidth}{@{\extracolsep{\fill}} lcccc @{}}
\toprule
& \multicolumn{2}{c}{\textbf{Depth}} & \multicolumn{2}{c}{\textbf{Point}} \\
\cmidrule(lr){2-3} \cmidrule(lr){4-5}
Method & Rel$\downarrow$ & $\delta\uparrow$ & Rel$\downarrow$ & $\delta\uparrow$ \\
\midrule
No cond               & 4.04 & 96.8 & 5.55 & 95.6 \\
Ours Intrin cond      & 3.98 & 96.8 & 5.08 & 96.2 \\
Ours Sparse D cond    & 3.21 & 98.2 & 4.60 & 96.9 \\
Ours D + K            & 3.15 & 98.3 & 3.59 & 97.9 \\
\bottomrule
\end{tabular*}
\label{tab:ablation_cond_inputs}
\vspace{-2mm}
\end{table}

\subsection{Experimental Results and Analysis}

\begin{figure*}[t]
  \centering
  \includegraphics[width=\linewidth, trim={0mm 0cm 0 0cm},clip]{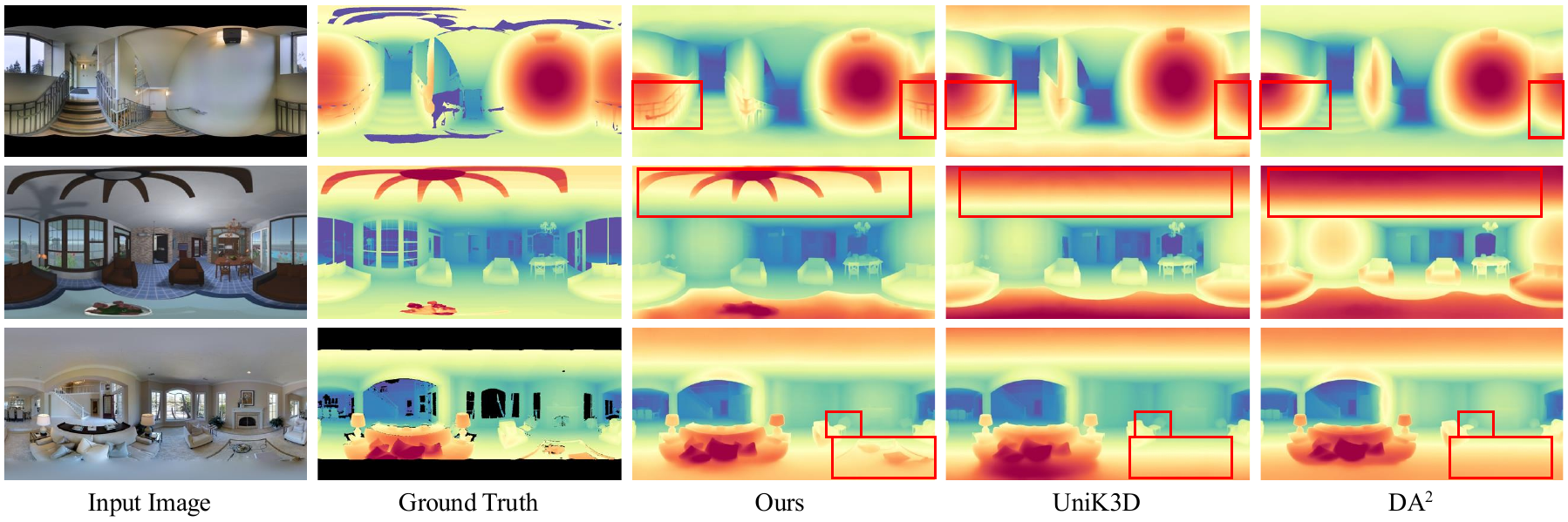}
  \caption{\textbf{Qualitative Comparison with Baselines on Panorama Image.} Our method produces sharper panorama depth maps by adding our bidirectional data augmentation strategy.}
  \label{fig:compare2}
\end{figure*}

\paragraph{Relative Geometry and Depth}
We evaluate the relative depth and geometry performance of \paper{} across different camera models, with comparisons shown in Tab.~\ref{tab:relative_geo}, Fig.~\ref{fig:compare1}, and Fig.~\ref{fig:compare2}.
On standard perspective images (S.FoV), our method achieves highly robust performance (96.8 $\delta_1$), proving that our universal formulation preserves accuracy in traditional settings, matching strictly pinhole-specialized models like MoGe V2. The true superiority of our approach emerges in non-pinhole environments. When applied to large-FoV fisheye images, standard pinhole models fail at point map prediction due to severe projection distortion. In contrast, \paper{} drastically surpasses all baselines, reducing the relative error of the universal baseline UniK3D from 11.5 down to 6.66 Rel. On $360^\circ$ panoramic images, \paper{} achieves the best depth accuracy (5.79 Rel), significantly outperforming the leading panorama-specific model DA$^2$ (6.65 Rel) while recovering sharper boundaries and undistorted structural lines, as visually evidenced in Fig.~\ref{fig:compare2}. The failure of UniK3D to maintain proper vertical structures on panoramas (Fig.~\ref{fig:compare1}) further validates the advantage of our bidirectional augmentation over purely latent unprojection representations.
These results highlight our model's unique capability to serve as a universal geometry estimator.

\paragraph{Metric depth}
Table~\ref{tab:metric_depth} compares metric-scale depth estimation on various benchmarks. In the zero-shot, image-only setting, our model (83.5 Mean) achieves state-of-the-art performance, surpassing previous best metric estimators like UniK3D (82.6 Mean) and MoGe V2 (77.5 Mean). This confirms that our universal ray-distance representation and diverse training data lead to robust metric-scale understanding. Furthermore, we show the benefit of our flexible conditioning. By providing camera intrinsics, performance consistently improves, validating our intrinsic injection module. When provided with sparse depth priors, the performance sees a dramatic boost (98.0 Mean), demonstrating our model's capability to effectively utilize and densify sparse geometric information, akin to a high-performance depth completion system.

\paragraph{Geometric Conditioning}
We provide an analysis of our geometric conditioning mechanism in Tab.~\ref{tab:ablation_cond_inputs} and Tab.~\ref{tab:metric_depth}. Starting from the unconditioned baseline (4.04 Depth Rel), injecting exact camera intrinsics immediately improves both depth (3.98 Rel) and point (5.08 Rel) accuracy by removing implicit unprojection ambiguities. When provided with sparse depth priors, the model seamlessly transitions into a high-performance densification system, yielding a massive boost across all metrics (\eg, zero-shot metric $\delta_1$ surges to an exceptional 97.5 Mean in Tab.~\ref{tab:metric_depth}). Finally, providing both priors simultaneously yields the highest precision (3.15 Depth Rel), confirming that our network efficiently fuses synergistic geometric information without suffering from feature distribution shifts.

\begin{table}[t]
\centering
\begin{minipage}{0.48\columnwidth}
\centering
\caption{\textbf{Output Representations.} Our decoupled Ray+D representation consistently improves the performance across all camera types.}
\begin{tabular}{lcccccc}
\toprule
& \multicolumn{2}{c}{\textbf{S.FoV}} 
& \multicolumn{2}{c}{\textbf{L.FoV}} 
& \multicolumn{2}{c}{\textbf{360 Images}} \\
\cmidrule(lr){2-3} \cmidrule(lr){4-5} \cmidrule(lr){6-7}
Method & Rel$\downarrow$ & $\delta\uparrow$ 
       & Rel$\downarrow$ & $\delta\uparrow$
       & Rel$\downarrow$ & $\delta\uparrow$ \\
\midrule
Ray+D & \textbf{4.04} & \textbf{96.8} & \textbf{6.37} & \textbf{94.0} & \textbf{5.79} & \textbf{95.1} \\
XYZ   & 4.28 & 96.3 & 6.74 & 93.8 & 6.28 & 94.5 \\

\bottomrule
\end{tabular}
\label{tab:ablation_output_rep}
\end{minipage}
\hfill
\begin{minipage}{0.48\columnwidth}
\centering
\caption{\textbf{Point Loss Mechanisms.} Our decoupled loss (P + GT Ray) isolates distance errors for the highest accuracy.}
\begin{tabular}{lcccc}
\toprule
& \multicolumn{2}{c}{\textbf{Depth}} 
& \multicolumn{2}{c}{\textbf{Point}} \\
\cmidrule(lr){2-3} \cmidrule(lr){4-5}
Method & Rel$\downarrow$ & $\delta\uparrow$ 
       & Rel$\downarrow$ & $\delta\uparrow$ \\
\midrule
P + GT Ray & \textbf{4.04} & \textbf{96.8} & \textbf{5.55} & \textbf{95.6} \\
P          & 4.05 & 96.8 & 5.65 & 95.2 \\
D + Ray    & 5.01 & 96.4 & 7.88 & 93.5 \\
\bottomrule
\end{tabular}
\label{tab:ablation_loss}
\end{minipage}
\end{table}

\begin{table}[t]
\centering
\begin{minipage}{0.48\columnwidth}
\centering
\caption{\textbf{Bidirectional Augmentation.} Perspective-to-Any (P2A) synthesis and Any-to-Perspective (A2P) self-training both independently improve panoramic predictions. Used synergistically, they effectively overcome the extreme scarcity of non-pinhole 3D data.}
\begin{tabular}{cccc}
\toprule
P2A & A2P & Rel$\downarrow$ & $\delta\uparrow$ \\
\midrule
-- & -- & 8.45 & 89.8 \\
\checkmark & -- & 6.21 & 94.5 \\
-- & \checkmark & 8.21 & 90.9 \\
\checkmark & \checkmark & \textbf{5.79} & \textbf{95.1} \\
\bottomrule
\end{tabular}
\label{tab:ablation_aug}
\end{minipage}
\hfill
\begin{minipage}{0.48\columnwidth}
\centering
\caption{\textbf{Conditioning Mechanisms.} Removing Gaussian smoothing severely degrades sparse-guided performance by injecting irregular noise, while removing the state embedding causes feature distribution shifts. Both components are vital for robust prior injection.}
\begin{tabular}{lcc}
\toprule
Method & Rel$\downarrow$ & $\delta\uparrow$ \\
\midrule
Ours            & \textbf{3.21} & \textbf{98.2} \\
w/o Smooth      & 3.80 & 97.5 \\
w/o State Emb   & 3.38 & 97.7 \\
\bottomrule
\end{tabular}
\label{tab:ablation_cond_design}
\end{minipage}
\vspace{+2mm}
\end{table}

\subsection{Ablation Studies}
\vspace{-2mm}

\paragraph{Output Representations}
Table~\ref{tab:ablation_output_rep} validates our ray-distance representation against a baseline that directly predicts global $(x,y,z)$ coordinates. As shown, the XYZ formulation consistently lags behind our decoupled approach across all camera types, including standard perspective (4.04 vs 4.28 Rel), fisheye (6.37 vs 6.74 Rel), and panoramic images (5.79 vs 6.28 Rel). This explicitly confirms our core hypothesis: forcing a single neural network to implicitly memorize and alternate between diverse, complex unprojection mappings creates severe optimization conflicts. By explicitly decoupling projection geometry from scene structure, our Ray+D representation resolves these conflicts, enabling superior generalization across all camera models.

\paragraph{Decoupled Loss}
We validate our decoupled point loss (Sec.~\ref{sec:decoupled_loss}) in Tab.~\ref{tab:ablation_loss}. Independent, disjoint supervision of distance and ray (D + Ray) completely fails to enforce spatial geometric consistency, resulting in poor point cloud accuracy (7.88 Rel). Conversely, a naive entangled point loss (P) allows gradient cross-contamination, where unprojection errors inappropriately penalize structural predictions, visibly degrading point quality compared to our method (5.65 vs 5.55 Rel). By projecting the predicted distance strictly along the ground-truth ray (P + GT Ray), our loss safely isolates structural errors, yielding the most stable optimization target.

\paragraph{Bidirectional Augmentation}
We ablate our data augmentation strategy for $360^\circ$ images in Tab.~\ref{tab:ablation_aug}. Without any augmentation, the model heavily overfits to the limited available data, resulting in poor panoramic geometry (8.45 Rel). Our Perspective-to-Any synthesis generates crucial synthetic supervision, providing the single largest performance jump (to 6.21 Rel). Any-to-Perspective self-training provides highly complementary pseudo-supervision directly from real-world omnidirectional scenes. Combining both techniques synergistically yields our final state-of-the-art result of 5.79 Rel.

\paragraph{Condition Design}
Table~\ref{tab:ablation_cond_design} validates the specific components of our geometric conditioning module (Sec.~\ref{sec:geo_condition}). Removing the Gaussian splatting procedure severely degrades sparse-guided performance (from 3.21 to 3.80 Rel), proving that raw sparse points are too irregular to provide stable gradients and must be normalized into dense, spatially consistent guidance. Furthermore, removing the learnable state embeddings causes a distinct performance drop (3.38 Rel). This confirms that explicitly signaling the active priors is strictly necessary to resolve architectural ambiguity and prevent harmful feature distribution shifts within the ViT backbone.

\section{Conclusion}
We introduce \paper{}, a unified framework for monocular geometry estimation that jointly handles arbitrary camera models and diverse input priors. The core of our method is a universal ray-distance representation that fundamentally decouples projection geometry from structural geometry. To overcome data scarcity, we proposed a systematic bidirectional data augmentation pipeline bridging perspective and omnidirectional imagery. Finally, our robust prior-injection mechanism seamlessly integrates sparse depth and camera intrinsics when available. Extensive experiments demonstrate that \paper{} achieves state-of-the-art zero-shot performance across pinhole, fisheye, and panoramic domains, establishing a robust new standard for 3D vision.

\bibliographystyle{splncs04}
\bibliography{main}

\clearpage
\setcounter{page}{1}
\appendix

\section{Training Datasets}
\label{sec:training_dataset}
We utilize 29 labeled datasets with ground truth depth annotation, together with two unlabeled panoramic datasets for geometry self-training.
Tab.~\ref{tab:dataset_full} provides a detailed taxonomy of all data sources.
Our dataset design follows three principles:

\begin{itemize}
    \item \textbf{Camera diversity.}
    We include standard pinhole images for strong baseline geometry, fisheye images common in autonomous driving systems, and full 360$^\circ$ panoramas used in AR/VR and embodied perception.
    
    \item \textbf{Scene diversity.}
    Our training data spans indoor, outdoor, mixed, and object-centric scenes, ensuring robustness to different spatial scales, environments, and viewpoint patterns.

    \item \textbf{Real + Synthetic Complementarity.}
    Synthetic datasets contribute clean ground-truth geometry, while real data reduces domain gaps and enhances in-the-wild generalization.
\end{itemize}

To mitigate the scarcity of fisheye and panoramic supervision, we apply the bidirectional data augmentation introduced in the main paper.
Perspective-to-Any converts existing pinhole datasets into synthetic wide-FoV views with accurate ground truth.
Any-to-Perspective leverages abundant unlabeled panoramic videos to refine wide-FoV performance via pseudo-labeling.
This dataset strategy is critical to closing the performance gap between pinhole and non-pinhole inputs and enables \paper{} to operate reliably on any camera model in the wild.
It is worth noting that to prevent data leakage in fisheye images, we strictly split the train and test sets by sequence.

\begin{table*}[t!]
\centering
\caption{\textbf{Training Datasets.} 
We train \paper{} on a highly diverse collection of 29 labeled datasets and 2 unlabeled panoramic datasets, spanning three major camera models: pinhole, fisheye, and 360° panoramic. 
These datasets cover indoor, outdoor, and mixed scene types, as well as both object-centric and large-scale environments, with both real and synthetic sources. 
}
\resizebox{0.95\linewidth}{!}{
\begin{tabular}{llccc}
\toprule
\textbf{Camera Type} & \textbf{Dataset Name} & \textbf{Scene Type} & \textbf{Metric?} & \textbf{Real?} \\
\midrule
\multirow{25}{*}{Pinhole} 
& 3D Ken Burns~\cite{3dkb} & Mixed & No & Synthetic \\
& ARKitScenes~\cite{arkitscenes} & Indoor & Yes & Real \\
& ARKitScenes-HighRes~\cite{arkitscenes} & Indoor & Yes & Real \\
& BEDLAM~\cite{bedlam} & Mixed & Yes & Synthetic \\
& BlendedMVS~\cite{blendedmvs} & Mixed & No & Synthetic \\
& Dynamic Replica~\cite{dynamicReplica} & Indoor & Yes & Synthetic \\
& EDEN~\cite{eden} & Outdoor & Yes & Synthetic \\
& HOI4D~\cite{hoi4d} & Indoor & Yes & Real \\
& HyperSim~\cite{hypersim} & Indoor & Yes & Synthetic \\
& IRS~\cite{irs} & Indoor & Yes & Synthetic \\
& Matterport3D~\cite{matterport3d} & Indoor & Yes & Real \\
& MVS-Synth~\cite{mvssynth} & Outdoor & Yes & Synthetic \\
& OmniObject3D~\cite{omniobject3d} & Object-Centric & Yes & Synthetic \\
& OmniWorld-Game~\cite{omniworld} & Mixed & No & Synthetic \\
& PointOdyssey~\cite{pointodyssey} & Mixed & Yes & Synthetic \\
& ScanNet++~\cite{scannetpp} & Indoor & Yes & Real \\
& SmartPortraits~\cite{smartportraits} & Indoor & Yes & Real \\
& Spring~\cite{spring} & Mixed & Yes & Synthetic \\
& Synscapes~\cite{synscapes} & Outdoor & Yes & Synthetic \\
& TartanAir~\cite{tartanair} & Mixed & Yes & Synthetic \\
& TartanAir v2~\cite{tartanair} & Mixed & Yes & Synthetic \\
& Unreal4K~\cite{unreal4k} & Mixed & Yes & Synthetic \\
& UrbanSyn~\cite{urbanSyn} & Outdoor & Yes & Synthetic \\
& VirtualKITTI2~\cite{vkitti} & Outdoor & Yes & Synthetic \\
& Waymo~\cite{waymo} & Outdoor & Yes & Real \\
& WildRGBD~\cite{wildrgbd} & Object-Centric & Yes & Real \\
\midrule
\multirow{1}{*}{Fisheye}
& KITTI360~\cite{kitti360} & Outdoor & Yes & Real \\
\midrule
\multirow{4}{*}{Panorama}
& Structure3D~\cite{Structured3D} & Indoor & Yes & Synthetic \\
& Matterport3D~\cite{matterport3d} & Indoor & Yes & Real \\
& Diverse360~\cite{diverse360} & Mixed & -- & Real \\
& 360+x~\cite{chen2024360+} & Mixed & -- & Real \\
\bottomrule
\end{tabular}
}
\label{tab:dataset_full}
\end{table*}

\section{More Implementation Details}
\label{sec:impl_details_supp}

\paragraph{Backbone and Prediction Heads}
We adopt DINOv2 ViT-Large~\cite{dinov2} as the encoder with the patch size of 14.
Following MoGe V2~\cite{mogev2}, the decoder heads consist of lightweight convolutional branches:
(1) the \textit{ray} and \textit{distance} prediction heads follow a DPT-style design~\cite{dpt} for efficient dense regression,
(2) the \textit{mask} head shares the same structure, and
(3) the \textit{metric} head is a two-layer MLP that processes the learned metric token and outputs a global scale factor $m$.
This design keeps inference efficient while supporting unified geometry prediction.

\paragraph{Training Strategy}
Training is conducted in three stages:
\begin{itemize}
    \item \textbf{Stage 1: Perspective pretraining.}
    We initialize the backbone learning rate at $1{\times}10^{-4}$ and the heads at $1{\times}10^{-5}$, decaying both by a factor of 0.5 every 2{,}000 steps.
    \item \textbf{Stage 2: Camera-universal finetuning.}
    We set the learning rates to $5{\times}10^{-5}$ (backbone) and $5{\times}10^{-6}$ (heads), and train with our bidirectional augmentation. For the Perspective-to-Any augmentation, we apply the augmentation with a sampling ratio of $2\%$ per batch.
    \item \textbf{Stage 3: Mask-head refinement.}
    All model parameters except the mask head are frozen.
    We train the mask head for 500 steps using 4 A100 GPUs, batch size 32 per GPU, with learning rate $1{\times}10^{-5}$.
\end{itemize}

\paragraph{Geometric Conditioning Configuration}
Sparse depth is randomly sampled from ground-truth depth during training with a per-pixel sampling ratio of $0.05\%$--$0.1\%$, simulating lightweight depth sensors (e.g., low-beam LiDAR or sparse SfM).
Geometric conditions (intrinsics and sparse depth) are applied only in Stage 2 and Stage 3, and enabled with a probability of $90\%$ to preserve robustness under RGB-only inputs.
For evaluation in Tab. 3 and Tab. 4 of the main paper, we fix the number of sparse depth samples to 1{,}000 points per image to ensure consistent benchmarking.

\section{More Experimental Results}

\paragraph{Comparison with UniK3D}
In Tab.~\ref{tab:sh_comparison}, we compare our proposed geometric representation against the Spherical Harmonics (SH) based representation utilized by recent camera-universal works such as UniK3D~\cite{unik3d}. We evaluate across standard Small Field-of-View (S.FoV), Large Field-of-View (L.FoV), and full 360$^\circ$ images. While SH parameterization provides a continuous representation of the ray field, we find that it tends to introduce unwanted smoothing artifacts and struggles to capture abrupt geometric changes, particularly in highly distorted edge regions (as visually demonstrated in Fig. 4 and Fig. 5 of the main paper). Our simple, explicit ray-distance pairing consistently outperforms the SH-based approach across all camera types. It provides a more stable and high-fidelity target for network regression, which is especially evident in 360$^\circ$ images where our method reduces the relative error from 6.01 to 5.79.

\begin{table}[h]
\centering
\caption{Comparison of different ray field representations. We compare our explicit ray-distance pairing against the Spherical Harmonics (SH) based representation across different camera models.}
\label{tab:sh_comparison}
\begin{tabular}{l|cc|cc|cc}
\toprule
\multirow{2}{*}{Method} & \multicolumn{2}{c|}{S.FoV} & \multicolumn{2}{c|}{L.FoV} & \multicolumn{2}{c}{360 Images} \\
& Rel $\downarrow$ & $\delta \uparrow$ & Rel $\downarrow$ & $\delta \uparrow$ & Rel $\downarrow$ & $\delta \uparrow$ \\
\midrule
Ours & \textbf{4.04} & \textbf{96.8} & \textbf{6.37} & \textbf{94.0} & \textbf{5.79} & \textbf{95.1} \\
SH-based Representation & 4.37 & 96.6 & 6.85 & 93.7 & 6.01 & 93.2 \\

\bottomrule
\end{tabular}
\end{table}

\section{Limitations}
\label{sec:limitations}

Although \paper{} significantly improves the geometric sharpness of depth and point cloud predictions, particularly after synthetic data finetuning, it still has difficulty reconstructing extremely fine structures such as thin edges and hair. 
This limitation may be alleviated by stronger local feature aggregation or hybrid architectures designed for fine-detail recovery (e.g., Depth Pro~\cite{depth_pro}).

Additionally, despite incorporating unlabeled images during training, the scale of our wide-FoV data remains modest compared to the abundance of real in-the-wild fisheye and 360° imagery.
We plan to expand the dataset and explore large-scale, self-supervised learning objectives tailored for universal camera geometry, which we believe will further enhance performance in diverse real-world applications.

\end{document}